\documentclass{mva_style}
\usepackage{graphicx}
\usepackage{subcaption}
\usepackage{amssymb}
\usepackage{url}
\expandafter\def\expandafter\UrlBreaks\expandafter{\UrlBreaks
  \do\s\do\d}

\finalcopy 

\begin{document}
\title{Super accurate low latency object detection\\ on a surveillance UAV}

\urldef{\mailsa}\path|{maarten.vandersteegen, kristof.vanbeeck, toon.goedeme}@kuleuven.be|

\author{
   Maarten Vandersteegen, Kristof Van Beeck, Toon Goedem\'e \\
   EAVISE, KU Leuven - Campus De Nayer\\
   \mailsa
}

\maketitle

\section*{\centering Abstract}
{\itshape
Drones have proven to be useful in many industry segments such as security and surveillance, where e.g. on-board real-time object
tracking is a necessity for autonomous flying guards. Tracking and following suspicious objects is therefore required in real-time on
limited hardware. With an object detector in the loop, low latency becomes extremely important.
In this paper, we propose a solution to make object detection for UAVs both fast and super accurate. We propose a multi-dataset
learning strategy yielding top eye-sky object detection accuracy. Our model generalizes well on unseen data and can cope with different
flying heights, optically zoomed-in shots and different viewing angles. We apply optimization steps such that we achieve minimal
latency on embedded on-board hardware by fusing layers, quantizing calculations to 16-bit floats and 8-bit integers, with negligible loss
in accuracy. We validate on NVIDIA's Jetson TX2 and Jetson Xavier platforms where we achieve a speed-wise performance boost of
more than $10\times$.
}

\section{Introduction}
While UAVs are demonstrating their potential to offer support for numerous tasks in different industry sectors, there is a rising need
for automating their control. On-board vision sensors combined with detection and tracking algorithms are a promising source of
input for controlling sky-grade robots. In security and surveillance sectors for example, there is a need for detecting and tracking
objects on the ground from hovering vehicles with immediate feedback to steering controls. If one for instance uses the output
position of an object detector as input for a PID controller, the latency of the detector should be minimal. In these
scenarios, processing video feeds from an on-board camera remotely is hardly doable due to streaming latencies and non-deterministic
network throughput, such that on-board processing is highly recommended.
Existing algorithms already provide powerful object detection, but often perform poorly for either ground level objects seen from
higher altitudes or, if trained on a UAV dataset, perform poorly on frontal scenes. Next to that, state-of-the-art neural nets
require heavy processing power, making them unusable for on-board inference.

In this work we propose an accurate low-latency object detector for on-board processing on a surveillance UAV. The algorithm
detects common objects such as persons or cars and works well on unseen images from both frontal scenes and birds-eye views.
To the best of our knowledge, we are
first in achieving this due to the joint dataset training strategy where we train on a super set composed of a large generic common
object dataset (MS COCO \cite{lin2014}) and a smaller drone dataset (Visdrone2018 \cite{visdrone2018}).
We prove that this method is superior to training on a single dataset and even to transfer learning from one dataset (MS COCO)
to the other (Visdrone2018), leaving the latter approach behind by a margin of 3.5\% mAP on Visdrone2018 and even 50\% mAP on
MS COCO.

The target platforms are NVIDIA Jetson TX2 and NVIDIA Xavier. While previous works run embedded object detectors/trackers on
NVIDIA Jetson platforms \cite{tijtgat2017, han2015, blanco2018}, and stick to simple implementations, we go a number of steps further
in optimizing our network for the target platform. We fuse layers and quantize calculations from floats to half floats (TX2) and even to 8-
bit integers (Xavier). Our best model achieves a latency of 12ms which is more than $10\times$ faster than our baseline
detector on the same hardware.

We made our extensively optimized implementation, based on TensorRT, publicly available
\footnote{code: \url{https://gitlab.com/EAVISE/jetnet}}.

\section{Related work}

\begin{figure*}
    \centering
    \includegraphics[width=\textwidth]{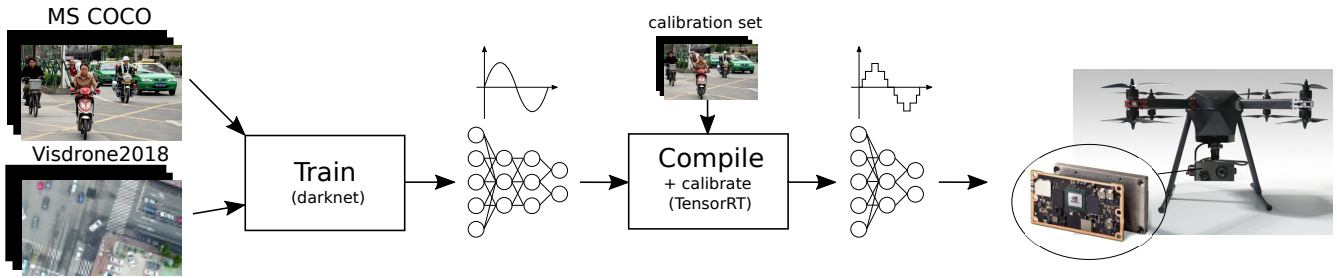}
    \caption{We train a model on MS COCO + Visdrone2018 and port the trained model to TensorRT to compile it to an inference
             engine which is executed on a TX2 or Xavier mounted on a UAV. During compilation, layer fusion and weight
             quantization is applied. A calibration set is used to restore the accuracy when using 8-bit integers. }
    \vspace{-1.5em}
    \label{fig:steps_taken}
\end{figure*}

As a first step, we start with the selection of the most suited object detector algorithm for our UAV application, used as a baseline.
Common network architectures for object detection include the (Fast(er)) R-CNN family \cite{ren2015} which go back to the very
beginning of the deep-learned object detection area and solve the problem in a two stage manner. Attempts to solve object detection
fast in one go result in SSD \cite{liu2016} and YOLO \cite{redmon2016}. Many more improvements were made to the fundamental proposals
from these detectors like the addition of Feature Pyramids \cite{lin2017}, the selection of better backbones like ResNet
\cite{he2016}, ResNeXT \cite{xie2017} and MobileNet \cite{howard2017}. These improvements result in networks as RetinaNet
\cite{lin2018}, Mask-RCNN \cite{he2017}, YOLOV3 \cite{redmon2018} and several other Faster-RCNN and SSD derivatives.
For our application we require an accurate object detector that can cope with both very small and large objects due to different flying
heights and zoom levels. The network architecture still needs to be fast due to the low latency criteria and the limited hardware
resources of our target platform. We selected YOLOV3 due to the excellent processing speed, the residual backbone for
efficient training and the feature pyramid to support the high dynamic range in object sizes.

Nowadays, many large scale datasets are available for training neural networks. For the sky-grade object detection and tracking
challenges for common objects in the wild, the most relevant available datasets are Visdrone2018 \cite{visdrone2018},
SDD \cite{sdd2016}, UAV123 \cite{uav123} and UAVDT \cite{uavdt}. Training an object detector to detect common objects in the wild is
a difficult task because it requires a large amount of high quality training data. Due to the limited viewpoints and natural variation
of most datasets, a network trained on dataset D is likely to perform well on D's test set but not necessarily on the test set of
dataset E. This so-called dataset bias \cite{torralba2011} causes difficulties in designing object detectors for real-life
applications. To overcome this problem we propose to train on a combination of a general and a task-specific dataset:
MS COCO \cite{lin2014} and Visdrone2018 \cite{visdrone2018}.
MS COCO is one of the few exceptions without a strong dataset bias and offers therefore the ideal training set
to combine with. Visdrone2018 adds support for bird-eye views and steep viewing angles and is chosen because of its similar relevant
class labels. In object detection, to the best of our knowledge, no other work explores joint training with multiple datasets.

Several works target embedded object detection or tracking on similar embedded hardware. Tijtgat et al. \cite{tijtgat2017}
compare YOLOV2 and tiny YOLOV2 among a few other classic object detection techniques both in terms of speed and accuracy on a Jetson TX2
platform for a UAV warning system. The same hardware is used by Blanco-Filgueira et al \cite{blanco2018}. They propose an embedded
real-time multi-object tracker based on a foreground-background detector and the GOTURN tracker. The predecessors of the TX2, the TX1
and TK1, are evaluated in \cite{han2015} where they combine a tiny version of Faster-RCNN with a KCF tracker for single-object
tracking on a drone. As stated before, we go further in optimizing our network for the target platform.
In addition, we make efforts to compress our trained network for faster inference.

Different techniques are
explored in literature like computation quantization from floats to half floats and 8-bit integers \cite{gao2018, tensorrt, gong2018},
network pruning \cite{he2018, molchanov2016, li2016}, architectural modifications \cite{howard2017, iandola2016} and network
distillation \cite{mehta2018}.
NVIDIA's TensorRT toolkit \cite{tensorrt} offers up to $4\times$ FLOPS reduction when quantizing the calculations from floats to 8-bit
integers thanks to mixed-precision GPUs. This however requires special training or calibration to cope with coarse quantization errors.
A state-of-the-art network pruning technique can suppress the number of calculations with an additional factor of up to $3\times$
\cite{molchanov2016} which is the case for simple networks like AlexNet and VGG. With more complex residual architectures,
the state-of-the-art reports a FLOPS reduction of up to 42\% without losing accuracy \cite{he2018}. Pruning however requires numerous cycles of fine-tuning
and increases the total training time with a large factor. One-time retrain pruning techniques \cite{li2016} also exist, but typically
result in much lower compression rates.
MobileNet and SqueezeNet \cite{howard2017, iandola2016} are conv-layer designs also targeted to reduce FLOPS and/or weights.
These modifications however, make it harder to train a network which usually results in lower precision.
Network distillation is also used as a compression technique to transfer the knowledge of a large network to a
smaller network. Mehta et al. \cite{mehta2018} report better performance with distillation compared to training a small network from
scratch. Although the distilled network's accuracy is 10\% mAP lower than its teacher, the speedup compared to the teacher is almost
$6\times$.

In our application, we strive for maximal reduction in FLOPS and minimal loss in accuracy. We therefore do not use MobileNet or
network distillation. Network quantization seems the most beneficial.

\section{Low latency object detection}
\label{sec:low_latency_object_detection}

In this section we discuss our methodology summarized in figure \ref{fig:steps_taken} and detail each of the steps:
how we design (\ref{sec:network_design}) and train (\ref{sec:training}) an algorithm based on YOLOV3 and how we optimize it
(\ref{sec:optimization}) for the target platform. We conclude this section by describing the hardware setup (\ref{sec:hardware_setup}).

\subsection{Network design}
\label{sec:network_design}
As stated earlier, we use YOLOV3 \cite{redmon2018} as our base architecture. Standard YOLOV3 is designed to detect 80 object classes
and supports a number of different square input resolutions (320x320, 416x416 or 608x608).

We make the following modifications:
\begin{enumerate}
    \item Select a custom set of supported object classes: \textit{person}, \textit{car}, \textit{bicycle}, \textit{motorbike},
    \textit{bus} and \textit{truck}
    \item Modify the input resolution to $608\times352$ (16/9 aspect ratio matching our camera)
    \item Recalculate the anchor boxes to best fit our training set
\end{enumerate}

The object class labels chosen are relevant to our application and all appear in the two datasets we use for training.
We reduce the number of output filters of the feature pyramid to support the six classes mentioned above.

\subsection{Training}
\label{sec:training}

\begin{figure}
    \centering
    \includegraphics[width=.35\textwidth]{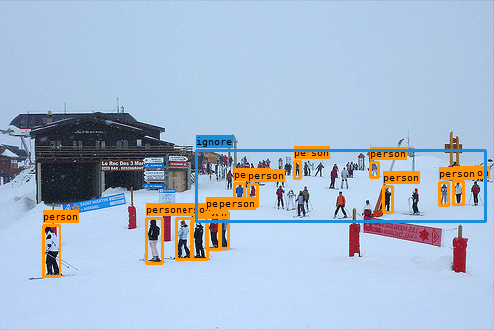} \\
    \vspace{0.5em}
    \includegraphics[width=.35\textwidth]{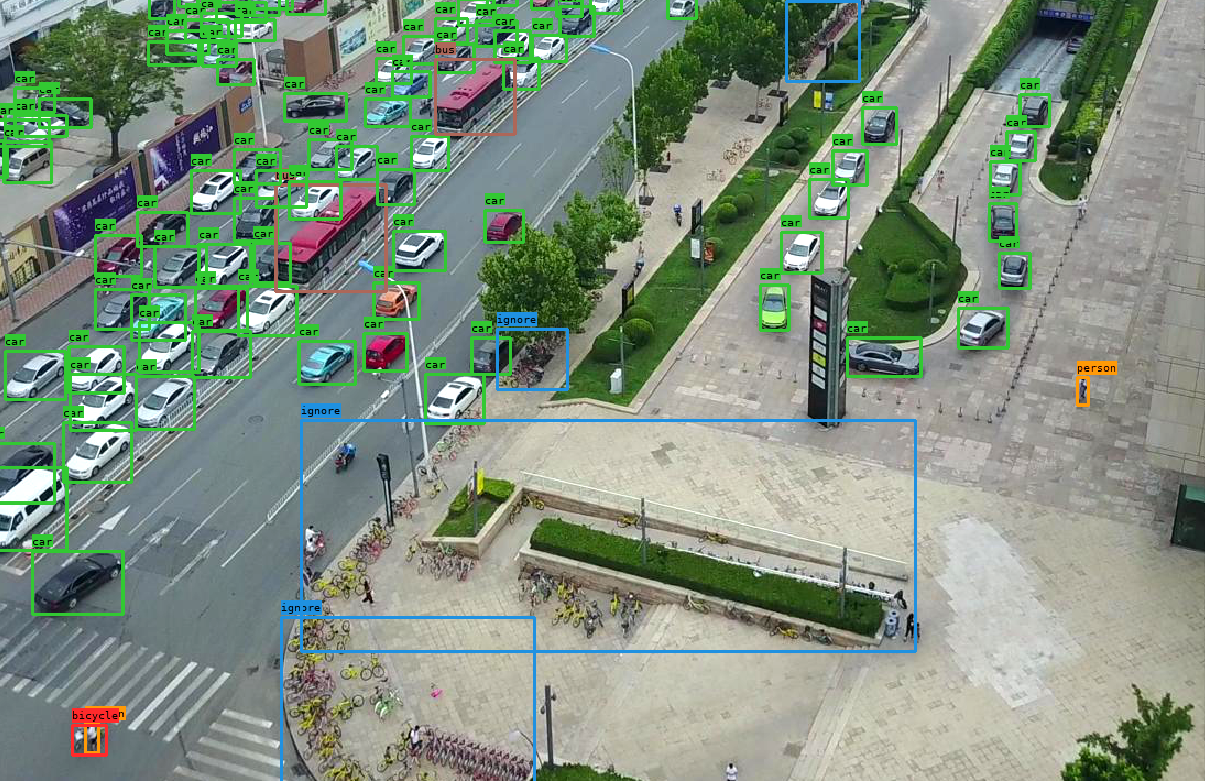}
    \caption{Examples from the MS COCO (top) and Visdrone2018 (bottom) training set, with their modified annotations.}
    \label{fig:training_samples}
\end{figure}

The detector is trained on a combination of two datasets: MS COCO \cite{lin2014} and Visdrone2018 \cite{visdrone2018}.
MS COCO contains 120k training images with
annotations for 80 different common object classes, mainly focusing on ground level scenes. Visdrone2018 has a training set of 6.5k
images and features 11 common object classes, seen from the sky from different angles and altitudes.
Also, both datasets contain images with dense cluttered objects which are bounded in rectangular regions, marked with the
\textit{iscrowd} flag (MS COCO) or with the \textit{ignored regions} class label (Visdrone2018).

For MS COCO, we remove all annotations not included in the supported class labels, listed in section \ref{sec:network_design},
leaving all resulting unannotated images in our training set as additional negatives \cite{torralba2011}.
Next, leftover annotations marked with the \textit{iscrowd} flag get a special \textit{ignore} label.
The loss function is modified to not penalize predictions inside these \textit{ignore} annotation areas as false positives during
training.

For Visdrone2018, we first remap all \textit{people} and \textit{pedestrian} labels to \textit{person} and all \textit{van} labels
to \textit{car} to match the supported class labels. Visdrone's labels \textit{ignored regions}, \textit{others} and
arguably \textit{tricycle} and \textit{awning-tricycle} are remapped to the special \textit{ignore} label. Figure
\ref{fig:training_samples} illustrates samples from the modified training data.

To show the added value of combining two datasets, we train three different models: a model only trained on Visdrone2018, a
transfer-learned model initialised with MS COCO weights and fine-tuned on Visdrone2018 and finally a model trained on both datasets.
For optimization purposes which are discussed in section \ref{sec:layerfusion}, we also train a ReLU variant of the latter model, where
all leaky ReLUs are replaced with standard ReLUs.

We use our own fork of the darknet framework \footnote{Training framework:
\url{https://gitlab.com/EAVISE/darknet}}
to train our models. Models trained or fine-tuned only on Visdrone2018
are trained for 100k iterations. Models trained on both datasets are trained for 500k iterations. Both are trained with a minibatch of 64
and a learning rate of 0.004 on four NVIDIA Tesla V100 GPUs. Training for 500k iterations takes up to 7 days.
Before training, all models are initialized with ImageNet classifier weights. For the ReLU model variant, we first train an ImagetNet
darknet53 classifier using ReLUs to get representative extraction weights for initialization of the ReLU model.
During training, the network input width
$w\in32\mathbb{N}:[416,960]$ is randomly scaled every 10 iterations together with the network input height $h\in32\mathbb{N}:[256,544]$
while trying to fit the $16/9$ aspect ratio as close as possible. Scaling the input resolution affects the scale of all layers.

\subsection{Optimization}
\label{sec:optimization}

It is previously demonstrated that network pruning of residual networks like YOLOV3 results in low compression rates \cite{li2016},
even with complex pruning algorithms \cite{he2018}. On top of that, a long iterative pruning process is needed to get a worthwhile
compression rate, specifically on large datasets. We therefore limit additional network optimization to layer fusion and calculation
quantization only.

\subsubsection{Porting to TensorRT}
\label{sec:porting}

Our framework is based on TensorRT \cite{tensorrt}, which we will use to accelerate our model for inference. The first step is to port
our trained model to TensorRT, where it will be compiled to a fast inference engine for our specific hardware target
(figure \ref{fig:steps_taken}). During compilation, layer fusion and calculation quantization are applied which are discussed in sections
\ref{sec:layerfusion} and \ref{sec:quantization}.

TensorRT is a very powerful inference engine but offers only a limited set of supported layer types. YOLOV3 contains two layer types
that are not supported: Leaky ReLU and the Upsample layer. The conventional way of dealing with unsupported layers in TensorRT is
creating a plugin layer with a custom implementation. We do this for both the Leaky ReLU and the Upsample layer.

\subsubsection{Layer fusion}
\label{sec:layerfusion}

When compiling the network to an inference engine, TensorRT first tries to fuse as many layers as possible. Typical convolution - batch
norm - activation blocks can be fused to a single layer, greatly improving inference speed. However, plugin layers do not allow to be
fused with other layers. Since our Leaky ReLU is a plugin,
it will not be fused with its proceeding batch norm/convolution layer. YOLOV3 contains 72
leaky ReLUs which may lead to a significant performance drop compared to using classic ReLUs.
As an alternative we therefore create a second implementation replacing the leaky ReLUs with ReLUs. This however requires retraining of
the network and might result in lower accuracy \cite{bing2015}.

\subsubsection{Quantization}
\label{sec:quantization}

\begin{figure}
    \centering
    \includegraphics[width=.3\textwidth]{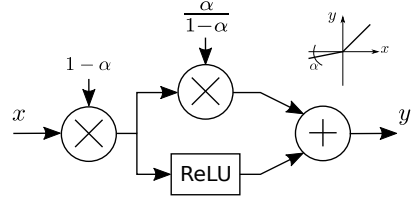}
    \caption{Leaky ReLU with negative slope $\alpha$ implemented with two scale layers, a ReLU layer and an element-wise adder.
             The first scale layer is likely to be fused into the previous layer.}
    \label{fig:native_leaky_relu}
    \vspace{-0.6em}
\end{figure}

Next to layer fusion, one can choose to quantize layer calculations to half floats and 8-bit integers.
All native layers can be quantised. Plugin layers however will still be executed using floats regardless of the quantization level
which forces TensorRT to insert type conversions before and after every plugin layer.
As stated in section \ref{sec:layerfusion}, our Leaky ReLU plugin is repeated 72 times which introduces 144 tensor type conversions.
This becomes even more problematic than the missed fusion opportunities. The alternative of replacing all leaky ReLUs with ReLUs
again solves this problem, but as stated before might not be as accurate. We propose a second alternative to
implement a Leaky ReLU with four natively supported layers as depicted in Figure \ref{fig:native_leaky_relu}. This implementation is
slower than a plugin implementation when using floats, but allows to be quantized which may result in better performance in half
float and 8-bit integer mode compared to its plugin alternative. Furthermore, this leaky ReLU can benefit slightly from layer fusion
since the first scale layer is likely to be fused to the previous layer (still leaving three unfused layers).

No additional precautions need to be made when converting the model to half float (16-bit floats) precision. For 8-bit integers
however, special training \cite{gong2018} or weight calibration \cite{tensorrt} is required. In this work we focus on calibration
only, as depicted in figure \ref{fig:steps_taken}, since it is much faster. We use TensorRT's built-in entropy calibrator \cite{tensorrt}
which calculates upper and lower boundaries of feature map activations for each layer. These boundaries are determined by minimizing
the information loss between the quantized activation values and the original activation values
(based on the Kullback-Leibler divergence) when sending numerous images from the training set through the network.
Once the boundaries are calculated, the lower boundary will represent -128 while the upper boundary will represent 127.
All values outside the boundaries are clipped. The image set used in calibration is called the calibration set. We use a calibration set
of 1000 images, randomly sampled from our training set.

\subsection{Hardware setup}
\label{sec:hardware_setup}
Our main target platform is the NVIDIA Jetson TX2. The TX2 has an onboard Pascal GPU with 256 CUDA cores, which supports float and half
float calculations. We use TensorRT V4.0 on this platform. We also investigate performance on the Jetson Xavier, NVIDIA's latest
platform, featuring two TPUs and an onboard Volta GPU with 512 CUDA cores which supports floats, half floats and 8-bit integer calculations.
Our Xavier is running TensorRT V5.0 and we will only be using its GPU. All experiments are done with maximum compute power enabled on
both platforms.

\section{Results}
\label{sec:results}

\subsection{Model accuracy}

\begin{table}
\caption{Accuracy results of our models (\%mAP), validated on MS COCO and Visdrone2018 validation sets respectively.}
\label{tab:model_accuracies}
\centering
\begin{tabular}{l|c|c}
\hline
    model & MS COCO & Visdrone2018 \\
\hline
 coco\_608 &                 67.83\% & 24.94\% \\
 visdrone\_608 &              4.36\% & 41.45\% \\
 visdrone\_coco\_init\_608 & 10.40\% & 41.72\% \\
 joint\_608 &               60.56\% & 45.27\% \\
 joint\_960 &               63.48\% & 56.78\% \\
 joint\_relu\_608 &         62.06\% & 44.66\% \\
 joint\_relu\_960 &         63.91\% & 56.09\% \\
\hline
\end{tabular}
\end{table}

Our trained models are evaluated on the validation sets of MS COCO and Visdrone2018. The evaluation metric used is mean
average-precision for an IoU of 0.5.
Table \ref{tab:model_accuracies} presents the validation results. The name of each model represents the dataset its trained on while
the trailing number indicates the input width. The default YOLOV3 model, \textit{coco\_608}, is a little better than other models on
MS COCO, but performs poorly on Visdrone2018. The contrast is even larger for models \textit{visdrone\_608} and
\textit{visdrone\_coco\_init\_608} which do quite well on Visdrone2018 but fail completely on MS COCO.
This might indeed indicate that MS COCO has a weaker dataset bias compared to Visdrone2018. Model \textit{visdrone\_coco\_init\_608}
differs from \textit{visdrone\_608} in that it is transfer-learned from MS COCO to Visdrone2018 by initializing with weights from
\textit{coco\_608} prior to training on Visdrone2018.
We propose \textit{joint\_608} (trained jointly on both datasets) as the best alternative since it performs well on both validation
sets and even outperforms \textit{visdrone\_608} and \textit{visdrone\_coco\_init\_608} by at least 3.5\% on Visdrone2018 and
by more than 50\% on MS COCO. This proves that training on a joint set is superior compared to training on a single dataset and
even to transfer learning from one dataset to the other.

\begin{figure*}
    \centering
    \includegraphics[width=\textwidth]{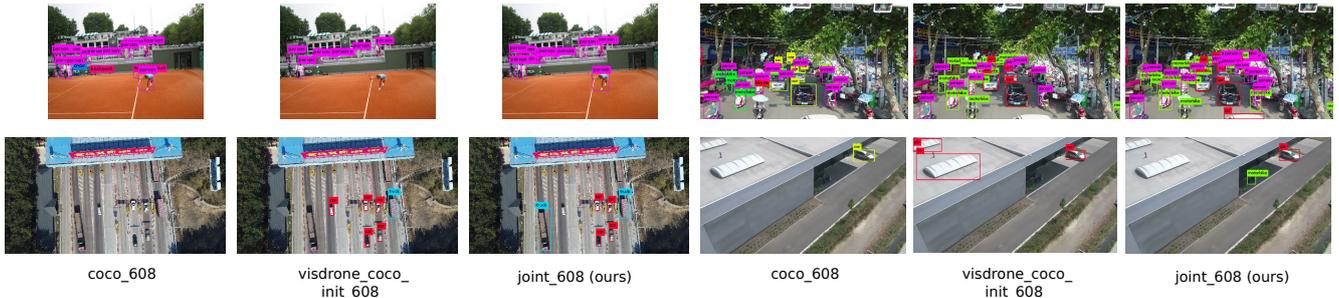}
    \vspace{-2em}
    \caption{Detection results of the \textit{coco\_608}, \textit{visdrone\_coco\_init\_608} and \textit{joint\_608} models on
             validation images of MS COCO (top left), Visdrone2018 (top right), UAVDT (bottom left) \cite{uavdt} and our own (bottom right).}
    \label{fig:detection_results}
\end{figure*}

We also tested \textit{joint\_960} which uses the same weights as \textit{joint\_608} but has an input resolution of $960\times544$.
This model outperforms \textit{joint\_608} by a large margin of 11\% for Visdrone2018, which can be explained by the fact that
Visdrone2018 contains very small instances which are better detected in a higher resolution. We only benchmark
the latency of \textit{joint\_960} on the Xavier platform since it is too heavy for the TX2.
As mentioned in section \ref{sec:training}, we also train a standard ReLU variant of our \textit{joint} model, coined
\textit{joint\_relu}. As expected, these models perform slightly worse (on Visdrone2018) compared to there leaky ReLU variants,
but the difference is less than 1\%. Figure \ref{fig:detection_results} gives a qualitative overview of the detection performance on
unseen data.

\begin{figure*}
    \centering
    \begin{subfigure}[b]{0.32\textwidth}
        \includegraphics[width=\textwidth]{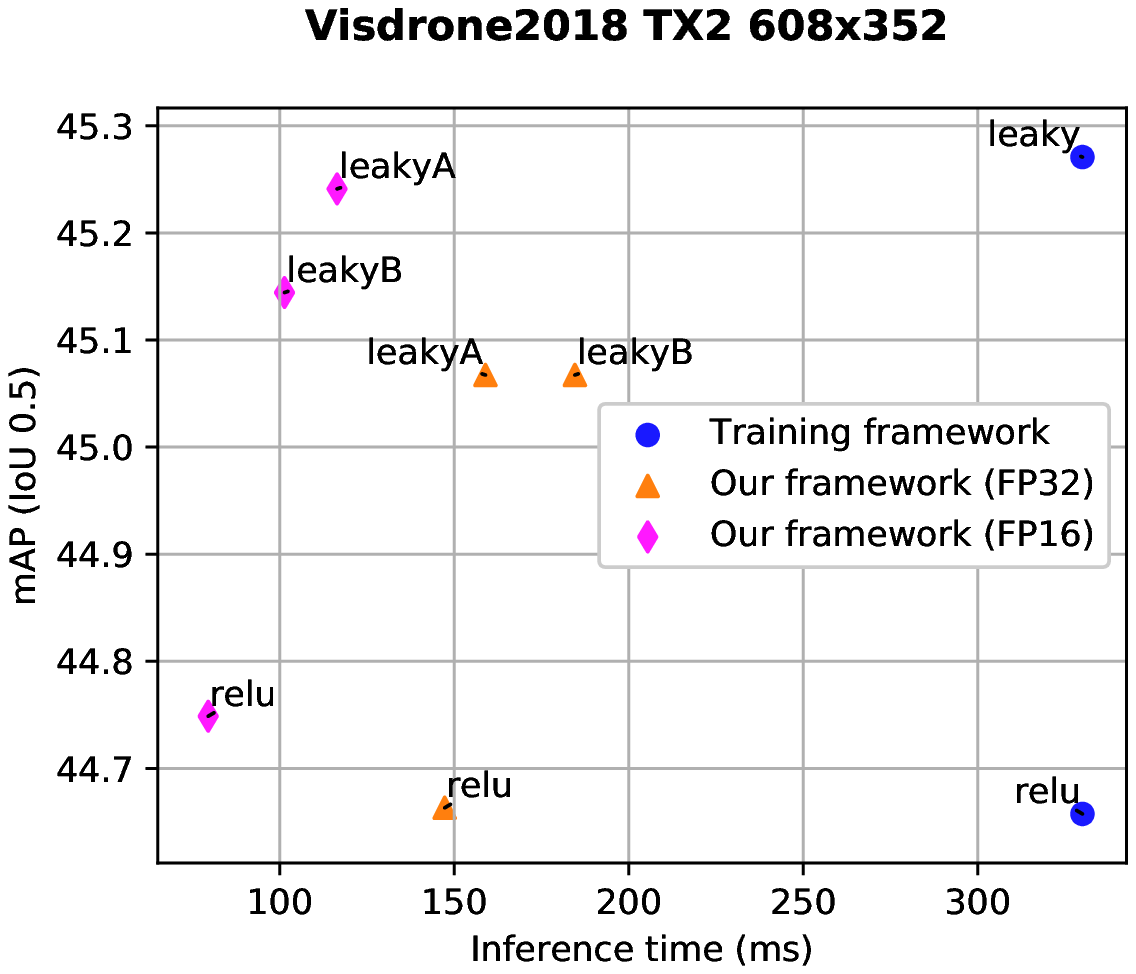}
        \label{fig:latecy_results_tx2}
    \end{subfigure}
    ~
    \begin{subfigure}[b]{0.32\textwidth}
        \includegraphics[width=\textwidth]{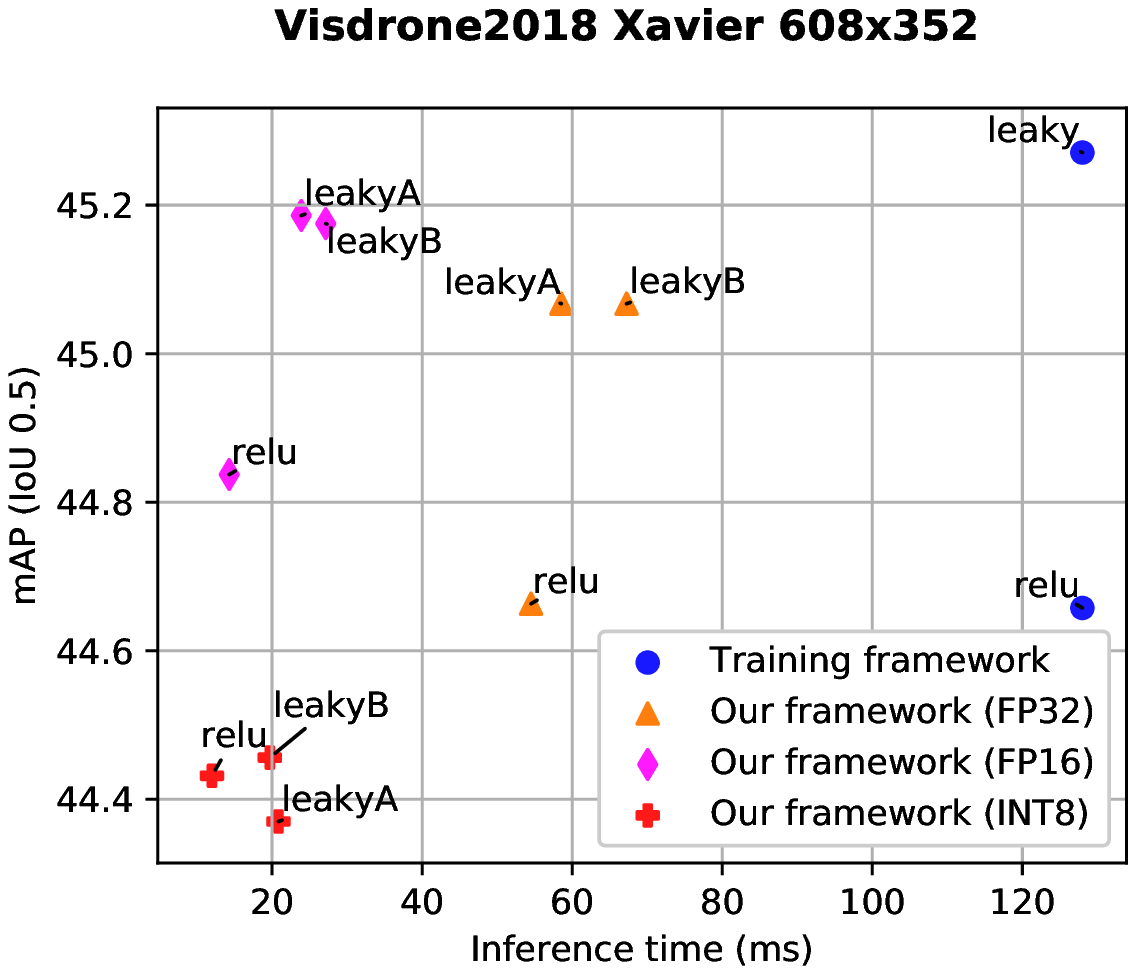}
        \label{fig:latecy_results_xavier_608}
    \end{subfigure}
    ~
    \begin{subfigure}[b]{0.32\textwidth}
        \includegraphics[width=\textwidth]{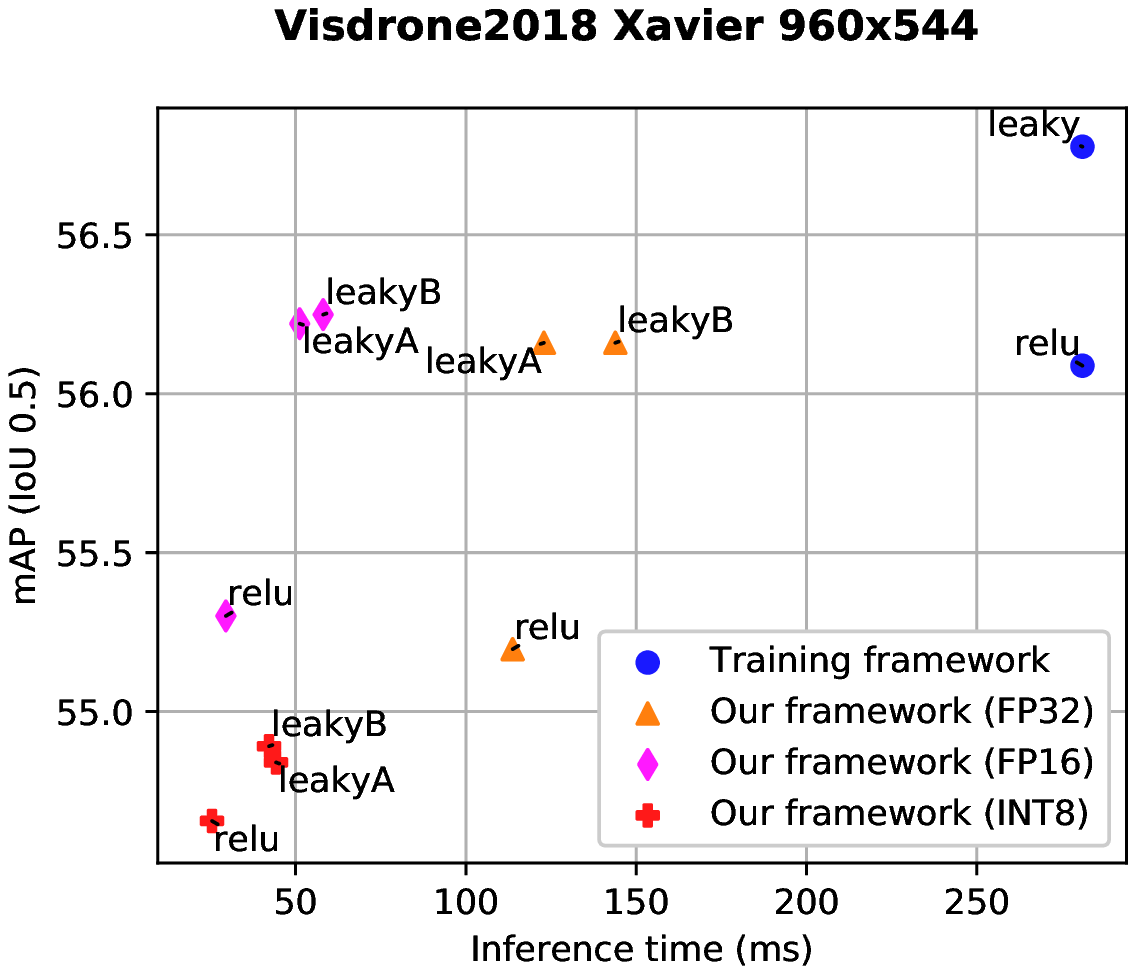}
        \label{fig:latecy_results_xavier_960}
    \end{subfigure}
    \vspace{-2em}
    \caption{Accuracy vs. latency results on the TX2 and Xavier platforms}
    \vspace{-1em}
    \label{fig:acc_vs_latency_results}
\end{figure*}

\subsection{Optimization}

Since we are interested in low-latency rather than throughput, we only benchmark our models with batch size equal to 1.
Figure \ref{fig:acc_vs_latency_results} presents our optimization results in accuracy (mAP) versus latency (ms) charts.
All measurements are GPU inference only (excluding pre- and postprocessing).
Label \textit{leakyA} represent optimized models using leaky ReLU plugin layers, while \textit{leakyB}
represents optimized models using our alternative leaky ReLU depicted in figure \ref{fig:native_leaky_relu}.
Benchmarks labeled with \textit{relu} represent optimizations with standard ReLU. The blue dots represent the baseline
(unoptimized) models, ran on the same hardware using our training framework with all GPU optimizations enabled.

On the TX2, the \textit{leakyB} optimization is 15\% faster than the \textit{leakyA} optimization in half float mode (FP16),
but still lacks behind our fastest TX2 model (\textit{relu} half float) which finishes in 79ms, outperforming its
baseline by $4.2\times$.

On the Xavier platform, \textit{leakyB} is only 6\% faster in 8-bit integer mode compared to the plugin version. The fastest model
finishes inference in about 12ms with a 608 input width and in 26ms with a 960 input width, boosting inference speed
$11\times$ compared to the baseline model.

Note that the accuracy drop of the optimized models compared to their baseline is negligible, even for the 8-bit
quantized models.

\section{Conclusion}
\label{sec:conclusion}

In this paper, we proposed a methodology to make object detection in UAV imagery both fast and super accurate. We achieved this by
introducing a multi-dataset learning strategy and demonstrated that this methodology significantly outperforms
transfer-learning by 3.5\% mAP on Visdrone2018 and by 50\% mAP on MS COCO, resulting in a generic model usable for all UAV
operations. Furthermore we applied extensive optimization steps to achieve minimal latency on our embedded target platforms,
the NVIDIA Jetson TX2 and Xavier. By fusing layers and quantizing calculations, we achieve a speed-wise performance
boost of $4.2\times$ on the TX2 and $11\times$ on the Xavier hardware without giving in on accuracy.

In future work, we plan to integrate our model with on-board trackers.

\section{Acknowledgement}

This work is partially supported by the 3DSafeGuard and Mirador project.

\end{document}